%% file: main.tex
\pdfoutput=1

\documentclass[10pt,twocolumn,letterpaper]{article}

\usepackage[pagenumbers]{wacv}

\input{preamble}

\definecolor{wacvblue}{rgb}{0.21,0.49,0.74}
\usepackage[pagebackref,breaklinks,colorlinks,allcolors=wacvblue]{hyperref}

\title{\pjn{}: Open-World Species Recognition on the Edge}

\author{Mohammad Mehdi Rastikerdar, Hui Guan, Deepak Ganesan\\
University of Massachusetts Amherst, Amherst, MA 01003, USA\\
{\tt\small \{mrastikerdar, huiguan, dganesan\}@cs.umass.edu}
}

\begin{document}
\maketitle

\input{sec/0_abstract}
\input{sec/1_intro}
\input{sec/2_related}
\input{sec/3_method}
\input{sec/4_experiments}
\input{sec/5_limitations}
\input{sec/6_conclusion}

{
    \small
    \bibliographystyle{ieeenat_fullname}
    \bibliography{bib}
}

\onecolumn
\appendix
\input{sec/7_supplementary}

\end{document}

%% file: preamble.tex
\usepackage{amsmath}
\usepackage{amssymb}
\usepackage{graphicx}
\usepackage{booktabs}
\usepackage{multirow}
\usepackage{subcaption}
\usepackage{xcolor}
\usepackage{enumitem}
\usepackage{array}

\newcommand{\pjn}{Scout}
\newcommand{\mypar}[1]{\vspace{0.05cm}\noindent{\textbf{#1.}\/}}

%% file: sec/0_abstract.tex
\begin{abstract}
Large vision-language models (VLMs) enable recognition beyond a fixed class set, but their computational demands prevent them from running on many edge devices. Cloud offload makes this capability accessible, but sending every image consumes scarce bandwidth and communication energy. We ask how to bring the open-world recognition capability of VLMs to the edge while operating within tight compute, energy, and bandwidth budgets. Wildlife monitoring provides a natural setting for exploring this question because camera traps encounter species not known at deployment. We present \pjn{}, an autonomous open-world recognition system that invokes a cloud VLM intermittently to teach new classes to a compact edge model. Given only the deployment location and empty site frames, \pjn{} autonomously turns each species identified by the VLM into persistent, site-conditioned recognition capability in a resource-efficient edge model, without a predefined species list, human labeling, or manual tuning. Across 30 camera-trap deployments in three regions on an NVIDIA Jetson Orin Nano, the accuracy of \pjn{} remains within 0.1--2.5\% of a model given a predefined species list. On species outside its initial class set, \pjn{} achieves 53.7--59.1\% accuracy, compared with 56.5--65.1\% for full cloud offload, while using 59--71\% less deployment energy.

\end{abstract}

%% file: sec/1_intro.tex
\section{Introduction}
\label{sec:intro}

\begin{figure*}[t]
    \centering
    \includegraphics[width=\textwidth]{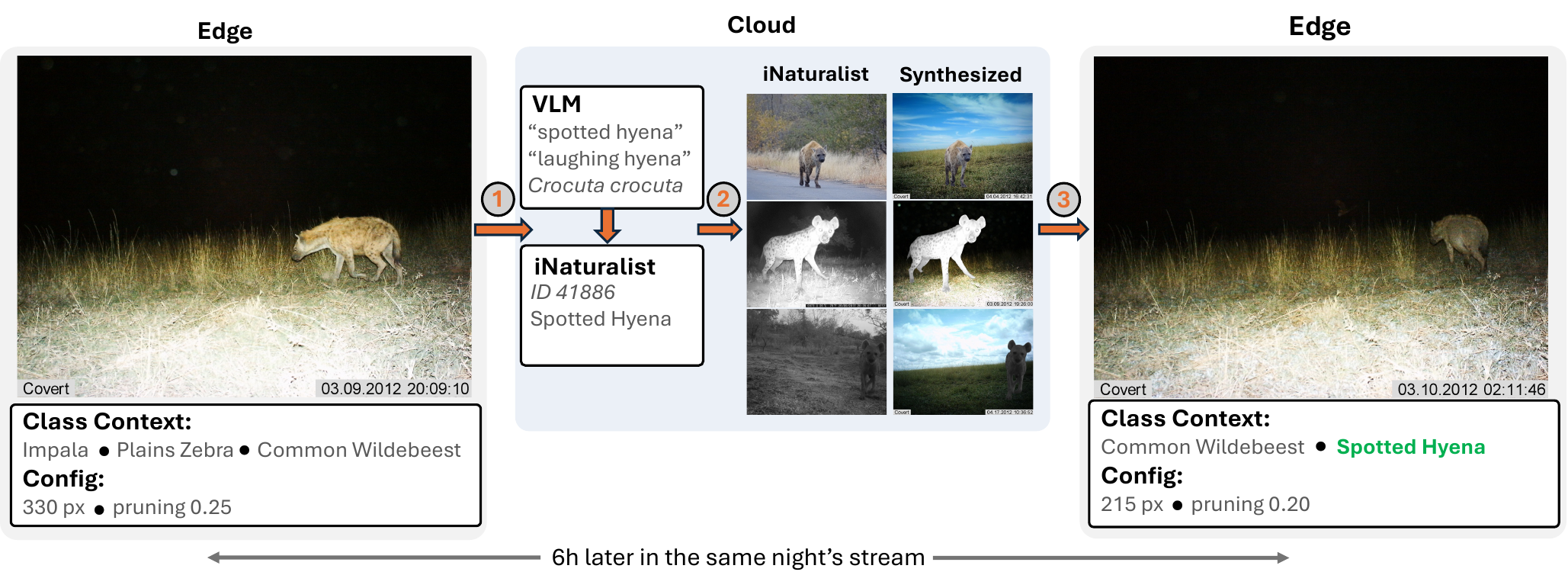}
   \caption{\pjn{} constructs site-conditioned
    training data and selects a resource-efficient model for the edge device. When
    the device encounters a species outside its current class set~(1), a cloud VLM
    names the species and \pjn{} automatically constructs new training data,
    retrains the classifier, and deploys the update~(2). The on-device classifier recognizes the same species six hours later in the same deployment stream~(3). Config lists the input image resolution and the model pruning ratio for the on-device classifier.}
    \label{fig:teaser}
\end{figure*}

Visual recognition is moving from a closed set, where all categories are fixed during training, toward an open world in which new categories can appear after deployment. Large vision-language models (VLMs) make this possible by recognizing categories not enumerated for a particular task~\cite{stevens2024bioclip,gu2025bioclip2,wildclip,catalog,taxabind}. However, their computational demands prevent them from running on many edge devices.

Cloud offload makes this capability accessible, but sending every image is costly when bandwidth and device energy are limited. Compact models can run locally, but recognize only classes seen during training. This creates a central question for edge vision: how can we bring the open-world capability of VLMs to the edge while operating within tight compute, energy, and bandwidth budgets?

Wildlife monitoring provides a natural setting for this question. Camera traps operate unattended for months in remote habitats to measure species presence and behavior~\cite{glover2019camera,ahumada2020wildlife}. Limited connectivity and human access make cloud communication and intervention expensive, while both the species and visual conditions vary across sites. A camera should therefore be deployable without a predefined species list or labeled animal images from the site and should learn new species as they appear.

\mypar{Open-world Recognition on the Edge} Rather than choosing between a fixed edge model and full cloud offload, the edge and cloud should play complementary roles. A compact model should handle familiar classes locally, while a cloud VLM should be consulted only when the edge encounters something it cannot recognize. Crucially, each cloud response should do more than answer the current query: it should teach the edge model the new class so that later observations can be processed locally. The VLM thereby becomes an intermittent teacher rather than a permanent inference engine. We call this problem \emph{open-world recognition on the edge}.

To realize this vision, a system must meet three requirements. First, it must detect and name an unfamiliar class. Second, it must obtain site-appropriate labeled data with which to learn the class. Third, it must update the on-device model under resource constraints. Existing work addresses only pieces of this problem: open-set recognition detects inputs outside the known classes but does not name or learn them~\cite{scheirer2013openset,bendale2016openmax,maxsoft1,openset1}; class-incremental learning assumes labeled examples of each new class; and edge--cloud and context-aware systems assume that the class set and training data are fixed before deployment~\cite{edgeboost,efficientedge,rastikerdar2024cactus}. What is needed instead is an end-to-end system that uses the cloud to identify new classes, constructs the data needed to learn them, and updates the edge model.

Open-world recognition alone is not sufficient because species recognition is highly sensitive to the deployment site. SpeciesNet~\cite{speciesnet}, for example, achieves 93.7\% accuracy at the Serengeti sites represented in its training data, but only 69.2\% at the unseen Nkhotakota sites, even though every evaluated species is in its label set. Resource constraints make this sensitivity more acute: large models can represent variation across many species and sites, while compact edge models must focus their limited capacity on the species and visual conditions relevant to one deployment. WildFiT~\cite{wildfit} shows that site-conditioned training data can improve compact camera-trap classifiers, but assumes a predefined species list and labeled animal examples. An open-world edge system must construct such data without knowing in advance which species it will encounter.

\mypar{Our Approach} We present \pjn{}, which uses a cloud VLM as an intermittent teacher to build and maintain a compact edge classifier. Given only the deployment location and empty site frames, \pjn{} combines public animal images with site backgrounds to construct site-conditioned training data. An LLM planner selects the class context, input resolution, and pruning ratio for the device. During deployment, the edge model handles its current context locally and uploads only out-of-context frames. A cloud response can trigger a different context or identify a new species; in the latter case, \pjn{} constructs new training data, retrains the classifier, and deploys the update. The VLM identification thereby becomes persistent edge capability, without human labeling, species curation, or model tuning. Figure~\ref{fig:teaser} summarizes the workflow.

Our work makes four major contributions:
\begin{itemize}[leftmargin=1.2em,itemsep=1pt,topsep=2pt]

\item We formulate \emph{open-world recognition on the edge}, which couples new-class discovery, autonomous training-data construction, and on-device model updates with the costs of local inference and cloud communication.

\item We develop an autonomous, site-conditioned training-data pipeline that turns a VLM-provided species name into labeled training and validation images using public animal images and backgrounds from the deployment site. The same pipeline supports initialization and learning new species during deployment.

\item We develop an LLM-guided planner that jointly selects the class context, input resolution, and pruning ratio using visual similarity and measured feedback from trained candidates.

\item We build and evaluate the complete system across 30 camera locations in three geographic regions on a Jetson Orin Nano. \pjn{} improves accuracy by 5.6--7.6\% at matched energy or reduces deployment energy by 35.8--63.2\% at matched accuracy, while achieving 53.7--59.1\% accuracy on species that fixed-label edge models cannot predict.

\end{itemize}

%% file: sec/2_related.tex
\section{Related Work}
\label{sec:related}

\mypar{Camera-trap Species Recognition}
Camera-trap species recognition has traditionally relied on supervised models
trained with labeled images from one or more deployment
sites~\cite{tabak2019machine,beery2018recognition,iwildcam2022-fgvc9,yousif2019animal}. SpeciesNet~\cite{speciesnet} scales this to over 2,000 labels, but such classifiers
often lose accuracy at locations absent from their training
data~\cite{beery2018recognition,speciesnet2}. Vision-language models broaden
coverage: BioCLIP and BioCLIP~2 learn taxonomy-aware representations from large
biological image collections~\cite{stevens2024bioclip,gu2025bioclip2}, and
TaxaBind extends this representation to location, satellite imagery, and
audio~\cite{taxabind}. Two models target camera traps directly: WildCLIP
retrieves images by natural-language description of the scene and the
animal~\cite{wildclip}, and CATALOG improves recognition when training and test
images come from different species or different sites~\cite{catalog}. These
methods still require a provided label set for classification and do not address
how a resource-constrained classifier is prepared and maintained for an
individual deployment.

\mypar{Site Adaptation and Data Synthesis}
Domain adaptation and generalization address accuracy loss when training and
deployment images are collected under different
conditions~\cite{motiian2017unified,peng2018zero,blanchard2011generalizing}.
This matters for camera traps because vegetation, illumination, weather, and
camera placement vary between sites~\cite{beery2018recognition} and drift over
time~\cite{camera_continual, wildfit}. Data synthesis reduces the domain gap by placing
foreground objects in target-domain
backgrounds~\cite{objectstitch,controlcom, AnyDoor}. WildFiT~\cite{wildfit}
applies this to camera traps, compositing labeled animal cutouts onto empty
frames from the deployment site and retraining when conditions change. \pjn{}
uses this composition procedure, but obtains the species list and reference
images from the deployment location and iNaturalist rather than requiring them as inputs.

\mypar{Open-Set and Continual Learning}
Open-set recognition detects images outside the training classes using signals such as softmax confidence, energy, or learned reciprocal
points~\cite{scheirer2013openset,bendale2016openmax,maxsoft1,
liu2020energyood,openset1}. Open-world recognition~\cite{bendale2015openworld} adds a loop that flags novel
classes and extends the classifier once they are labeled, and class-incremental methods add classes while reducing forgetting of earlier
ones~\cite{rebuffi2017icarl,kirkpatrick2017ewc,wildlife_incremental}. These
methods typically assume labeled examples of each new class are available. Our setting additionally requires naming a novel species and obtaining its training images without human annotation.

\mypar{Edge Inference and Model Selection}
Model compression and hardware-aware optimization reduce the cost of on-device inference~\cite{cai2020ofa,hinton2015distilling,jacob2018quantization,li2017pruning,tan2019mnasnet}. Edge--cloud systems cut communication by sending only low-confidence or difficult inputs to a cloud model~\cite{edgeboost,efficientedge}, and context-aware systems such as CACTUS and Palleon switch between models specialized for different class sets~\cite{rastikerdar2024cactus,palleon}. All of these fix the classes and training data before deployment. LLMs have also been used as optimizers, proposing candidates from a record of previously evaluated ones~\cite{opro,evoprompting}. \pjn{} lets the class set grow during deployment and uses an LLM to guide the search over class contexts and model configurations, training and evaluating every candidate before deploying it.

%% file: sec/3_method.tex
\section{Method}
\label{sec:method}

\begin{figure*}[t]
    \centering
    \includegraphics[width=\textwidth]{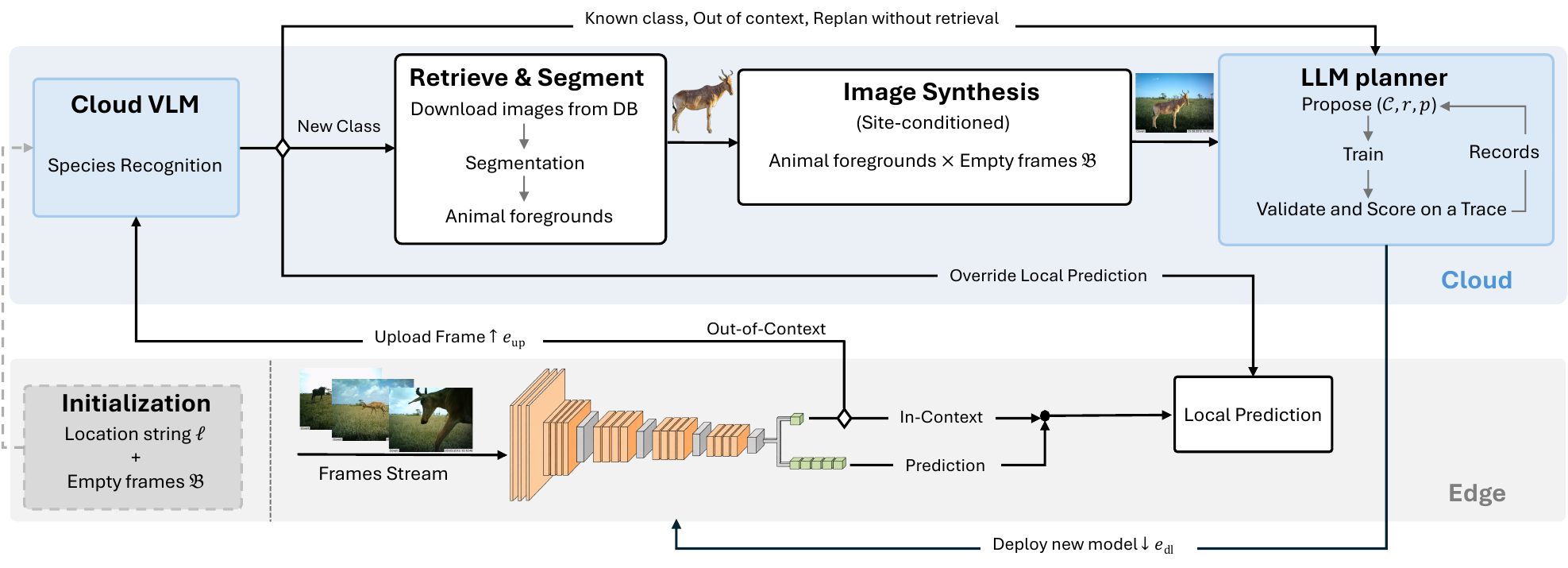}
    \caption{\textbf{\pjn{} system overview.} At initialization, \pjn{} identifies species likely to appear at the deployment location and constructs site-conditioned training data by combining animal foregrounds from iNaturalist with backgrounds from the deployment site. The LLM planner trains and evaluates candidate class contexts and model configurations before deploying a compact edge model. During deployment, the edge model handles familiar species locally and uploads only out-of-context (OoC) frames to the cloud VLM. A cloud response may correct an edge prediction, trigger a different class context, or identify a new species. For a new species, \pjn{} constructs additional training data and updates the edge model so that future observations can be processed locally.}
    \label{fig:system_overview}
\end{figure*}

\pjn{} uses a cloud VLM to autonomously build and maintain a compact classifier for an edge deployment. The user provides a location string $\ell$, a set of empty frames $\mathcal{B}$ captured at the site, and optionally a target accuracy $\alpha$. No species list or labeled animal images from the site are required. The cloud maintains a species pool $\mathcal{I}_t$, containing species that are expected at the site or have been identified during deployment, together with their training assets. At initialization, the location-conditioned VLM defines $\mathcal{I}_0$ by proposing species likely to appear at the site. The pool grows whenever the cloud VLM identifies a new species. The planner selects a smaller class context $\mathcal{C}_t \subseteq \mathcal{I}_t$ for the model currently deployed on the device. Figure~\ref{fig:system_overview} summarizes the complete workflow.

\subsection{Site-Conditioned Training Data Construction}
\label{sec:method:synthesis}

\pjn{} uses the same data-construction pipeline to build the initial classifier and to teach it each new species. At initialization, the VLM uses location $\ell$ to return scientific names of species likely to appear at the site. We request scientific names because common names can refer to different species in different regions. Each name is resolved to an iNaturalist species ID. If a name is missing or ambiguous, the VLM proposes an alternative and the lookup is repeated. A name that cannot be resolved is excluded.

For each resolved species, \pjn{} downloads research-grade images from iNaturalist. MegaDetector~\cite{megadetector}, a camera-trap animal detector, localizes the animal in each image, and SAM~\cite{kirillov2023segment} uses the detected bounding box to extract its foreground mask. The resulting foregrounds retain the species ID of the source image, which provides their label.

\pjn{} places these foregrounds on the provided empty frames from the deployment site using the composition procedure from WildFiT~\cite{wildfit}. Time matching pairs a foreground with a background captured at a similar time of day, herd-aware composition preserves groups of animals, and spatial preservation maintains the approximate position and scale of the animal in its source image. The resulting images combine species examples from iNaturalist with the background, illumination, and camera viewpoint of the deployment site.

Because \pjn{} records which species foreground it places in each image, the labels of the composed images are known rather than inferred. The same procedure therefore produces labeled training and validation images without labeled animal images from the deployment site. The downloaded images and extracted foregrounds are retained so that \pjn{} can train a different class context without repeating data acquisition.

\subsection{Cloud Teaching and Model Updates}
\label{sec:method:novel}

The on-device model contains a classification head and an out-of-context (OoC) head that share the same backbone. The classification head predicts among the species in the current context $\mathcal{C}_t$. Following CACTUS~\cite{rastikerdar2024cactus}, the OoC head is a regression head trained with label 0 using images of species in $\mathcal{C}_t$ and label 1 using images of expected or previously identified species in $\mathcal{I}_t \setminus \mathcal{C}_t$. Thus, OoC denotes a species outside the current deployed context, not necessarily outside the larger species pool. Both heads run in a single forward pass. An OoC score below $0.5$ returns the classification result locally, while a score of at least $0.5$ uploads the frame to the cloud VLM.

An OoC false positive uploads an in-context frame and consumes unnecessary communication energy. An OoC false negative classifies an unfamiliar species as one of the in-context species and prevents the VLM from teaching that species to the edge model. The OoC head therefore affects both recognition accuracy and the energy used for cloud consultation.

For an uploaded frame, the cloud VLM receives the image, deployment location, timestamp, and species pool. It is instructed to return a species ID from the pool when the animal matches a known species, or a scientific name when the species is absent from the pool. The response produces three cases. (1)~\textbf{In context.} The returned species belongs to $\mathcal{C}_t$. The cloud prediction replaces the on-device prediction, but the model is not updated. This case corresponds to an OoC false positive. (2)~\textbf{Outside context.} The returned species belongs to $\mathcal{I}_t$ but not to $\mathcal{C}_t$. The planner selects a new class context and model configuration using the training assets already stored for that species. (3)~\textbf{New species.} The returned name does not correspond to a species in $\mathcal{I}_t$. \pjn{} resolves the name to an iNaturalist species ID and applies the data-construction pipeline from Section~\ref{sec:method:synthesis}. The new species and its training assets are added to the species pool, after which the planner trains and selects an updated edge model. If the returned species cannot be resolved in iNaturalist, \pjn{} cannot construct training data and therefore does not update the edge model.

\subsection{LLM-Guided Model Optimization}
\label{sec:method:planner}

After initialization or a change in the species pool, \pjn{} must decide which recognition capability to place on the resource-constrained device. The planner jointly selects a class context $\mathcal{C}$ and model configuration $(r,p)$, where $r$ is the input resolution and $p$ is the structural pruning ratio. A larger class context handles more species locally and can reduce uploads, but may require greater model capacity to meet the target accuracy. Lower resolution and more pruning reduce inference energy, but may reduce classification accuracy or increase OoC errors and cloud communication.

The LLM serves as a proposal mechanism rather than an unverified decision maker. It proposes class contexts and model configurations using inter-species visual similarity and a planning record of previous measurements. \pjn{} trains and evaluates every proposed candidate; only a candidate with measured accuracy and an energy estimate can be selected for deployment.

\mypar{Device Calibration} At installation, the device creates variants of a profiling model using different pruning ratios and input resolutions. It measures their energy using on-device power sensors and fits a linear model relating model FLOPs to inference energy. The planner uses this relation to estimate the inference energy of candidates that have not yet been deployed.

The device also measures the energy required to upload the initial empty frames and download the profiling model. Dividing these measurements by the transferred data sizes gives the uplink and downlink energy per MB, denoted by $e_{\mathrm{up}}$ and $e_{\mathrm{dl}}$. These estimates are updated using later frame uploads and model downloads.

\mypar{Candidate Class Contexts} Because the edge model may not have sufficient capacity to cover the entire species pool, the planner considers class contexts of different sizes. It orders species by how recently they were observed and constructs nested candidates: a smaller context focuses the model on the most recently observed species, while larger contexts progressively include species observed earlier at the site. These candidates expose the trade-off between covering more species locally and concentrating the limited model capacity on fewer species.

To estimate how difficult each context will be for a compact model, the LLM uses a pairwise visual-similarity matrix over the species pool. Each entry is the cosine similarity between image embeddings from two species. Contexts containing visually similar species may require greater model capacity to distinguish them. The LLM combines this signal with the measured outcomes in the planning record to select a small set of contexts for evaluation. When a new species is added, its similarity values are initially unavailable. The LLM selects a visually related species in the pool as a proxy and temporarily uses its similarity profile.

\mypar{Candidate Training and Evaluation} For each selected context, the LLM proposes a model configuration, i.e., an input resolution and pruning ratio, using the visual-similarity matrix and the measured outcomes in the planning record. \pjn{} trains the corresponding classifier and OoC head using the synthesized images, then returns the candidate's measured accuracy and estimated energy to the LLM. If the candidate does not meet the target accuracy, the LLM proposes a configuration with greater capacity. Once the target is met, it searches for a lower-energy configuration that maintains the target. In this way, measured outcomes guide the search toward a model that provides sufficient accuracy without unnecessarily increasing resource use.

Each candidate is evaluated on a synthesized validation trace. During deployment, \pjn{} constructs the trace from the sequence of species predictions collected so far. For each entry, it creates an image using foregrounds of the corresponding species and backgrounds from the site. The trace therefore follows the estimated ordering and frequency of species at the deployment while providing recorded labels for candidate evaluation. Candidate accuracy uses the candidate prediction for frames handled locally and the cloud VLM prediction for frames triggered by the OoC head.

For a candidate model $m=(\mathcal{C},r,p)$ and a validation trace of $T_v$ images, the estimated deployment energy per image is

\begin{equation}
\begin{aligned}
\hat{E}(m) = \hat{E}_{\mathrm{inf}}(m)
+\frac{1}{T_v}\Big[
&e_{\mathrm{up}}\sum_j o_j S_{\mathrm{frame},j} \\
&+e_{\mathrm{dl}}\left(\bar{S}\sum_j u_j + S(m)\right)
\Big],
\end{aligned}
\label{eq:candidate_energy}
\end{equation}
where $\hat{E}_{\mathrm{inf}}(m)$ is estimated from model FLOPs, $o_j$ indicates that candidate $m$ uploads validation frame $j$, and $u_j$ indicates that the cloud response would trigger a later planning event and model update. $S_{\mathrm{frame},j}$ is the uploaded frame size, and $S(m)$ is the size of the current candidate. Because the planner does not know which model a future update will select, $\bar{S}$, the average size of previously downloaded models, approximates the size of each future update. The term $e_{\mathrm{dl}}S(m)$ includes the one-time cost of downloading the current candidate. Candidate energy is estimated because the model has not yet been deployed on the device.

\mypar{Planning Record and Final Selection} After every evaluation, \pjn{} stores the candidate context, input resolution, pruning ratio, measured accuracy, and estimated inference, upload, download, and total energy. Previously trained checkpoints are re-evaluated on the current validation trace using the latest communication costs, without retraining. The LLM receives this accumulated record when proposing candidates during later planning events.

When the user provides a target accuracy $\alpha$, \pjn{} selects the lowest-energy evaluated candidate that meets $\alpha$. If none reaches the target, it selects the candidate with the highest measured accuracy. Without a target accuracy, the LLM selects among the evaluated candidates according to the requested accuracy--energy trade-off.

%% file: sec/4_experiments.tex
\section{Experiments}
\label{sec:experiments}

\subsection{Experimental Setup}
\label{sec:exp:setup}

\mypar{Datasets}
We evaluate 30 deployments from three camera-trap datasets: Snapshot
Serengeti S04~\cite{snapshotserengeti}, Nkhotakota~\cite{nkhotakota}, and New
Hampshire Fish and Game Volume~1~\cite{nhdept,ammonitor}. From each dataset,
we select 10 camera locations containing at least 15 species after removing
unlabeled, non-species, and domestic-animal labels. We reserve 250 empty
frames from each location as the background pool $\mathcal{B}$ and process
the remaining images in timestamp order without using future observations.

\mypar{Evaluation setting}
We consider two initialization settings. In the \emph{controlled} setting,
all methods receive the 10 most frequent species in each dataset, while the
remaining species are not revealed. In the \emph{cold-start} setting,
\pjn{}-CS receives only the deployment location and $\mathcal{B}$, and the
VLM selects 10 initial species. We report overall accuracy and, in the
controlled setting, accuracy on species inside and outside the fixed initial
class set. For \pjn{}-CS, we report only overall accuracy because its initial
set is selected differently.

\mypar{Models and Training}
We use GPT-5~\cite{gpt5} as the cloud VLM and LLM planner. For each species,
we retrieve up to 250 iNaturalist images and use an 80/20 training-validation
split to construct the corresponding site-synthesized images. The on-device
models are structurally pruned, ImageNet-initialized EfficientNet-B0
models~\cite{efficientnet}; the planner selects an input resolution in
$[100,800]$ pixels and a pruning ratio in $[0,1)$. \pjn{} and the edge
baselines use the same synthesized data and configuration ranges whenever
applicable.

\mypar{Deployment Energy}
We report the average deployment energy consumed on the edge device per image:
\begin{equation}
E_{\mathrm{dep}} =
\frac{1}{T}\sum_{t=1}^{T}
\left[
E_{\mathrm{inf}}(m_t)
+ o_t E_{\mathrm{up},t}
+ u_t E_{\mathrm{dl}}(m_t^{+})
\right],
\label{eq:deployment_energy}
\end{equation}
where $m_t$ is the model used for image $t$, and $o_t$ and $u_t$ indicate an
image upload and model update, respectively. The three terms account for
on-device inference, uploading image $t$, and downloading the updated model
$m_t^{+}$.

We measure inference energy on a Jetson Orin Nano Super in
15\,W mode using \texttt{jtop}, after subtracting idle power. Reported
inference energy is measured for each deployed configuration; the
FLOPs-based model is used only to estimate candidate energy during planning. The datasets were collected at sites where we cannot directly measure the
communication link. We therefore estimate communication energy using the
transmit and receive power of a Quectel EG25-G cellular module~\cite{quectel} and regional throughput estimates from
SpeedChecker~\cite{speedchecker} and Opensignal~\cite{opensignal}. Exact
communication values and additional implementation details are provided in
the supplementary material (Section~\ref{sec:supp:experimental_details} and
Table~\ref{tab:supp:communication_energy}), and Section~\ref{sec:exp:sensitivity} evaluates
sensitivity to these estimates.

\mypar{Baselines}
We compare with three full-offload methods and six edge and edge--cloud configurations.
Full-Offload (GPT-5) sends every image to the same VLM used by \pjn{}.
SpeciesNet~\cite{speciesnet} uses MegaDetector~\cite{megadetector} followed
by EfficientNetV2-M, while BioCLIP~2~\cite{gu2025bioclip2} performs zero-shot
classification over all labels in its released TreeOfLife-200M species
vocabulary using the highest-confidence MegaDetector crop. \emph{On-Device-Only} runs an EfficientNet-B4 on the device.
\emph{EdgeBoost}~\cite{edgeboost} runs an EfficientNet-B0 on the device and
a B5 in the cloud. It accepts high-confidence local predictions
and offloads the remaining images using a threshold on the calibrated softmax
margin. \emph{EfficientEdge}~\cite{efficientedge} jointly trains an edge
classifier, a routing model, and an EfficientNet-B5 cloud classifier. We
evaluate it using B0 or B3 on
the edge. \emph{CACTUS}~\cite{rastikerdar2024cactus} uses a rule-based policy to switch among pruned B0 models trained for small class contexts, with a B5 model in the cloud. \emph{Palleon}~\cite{palleon} estimates the current class distribution
and switches among pruned B4 models. All edge
configurations retain their initial class set throughout deployment.

\subsection{End-to-End Performance}
\label{sec:exp:end_to_end}

Table~\ref{tab:end_to_end} reports accuracy and energy gain averaged over the
10 camera locations from each dataset. The three full-offload methods share
the same gain because each uploads every image and our deployment metric
excludes cloud inference energy. Computed deployment energy appears in
Figure~\ref{fig:accuracy_energy} and its breakdown into inference, upload,
and download in the supplementary material
(Table~\ref{tab:supp:end_to_end_complete}).

\begin{table}[t]
\centering
\caption{End-to-end performance averaged over 10 camera locations from each
dataset. Accuracy is reported overall and for known (K.) and novel (N.) species using the
partition of Section~\ref{sec:exp:setup}. Full-offload models do not use the
initial class set; but for comparison, K./N.\ reports their accuracy on the corresponding image
subsets. \pjn{}-CS selects its own initial set, so only its overall accuracy
is comparable. Energy gain is Full-Offload deployment energy divided by the
method's, so higher is better}
\label{tab:end_to_end}
\setlength{\tabcolsep}{2.0pt}
\renewcommand{\arraystretch}{1.12}
\resizebox{\columnwidth}{!}{%
\begin{tabular}{l cc cc cc}
\toprule
\multirow{2}{*}{Method}
& \multicolumn{2}{c}{Serengeti}
& \multicolumn{2}{c}{Nkhotakota}
& \multicolumn{2}{c}{New Hampshire} \\
\cmidrule(lr){2-3}
\cmidrule(lr){4-5}
\cmidrule(lr){6-7}
& Acc.\ (K./N.) & E. Gain
& Acc.\ (K./N.) & E. Gain
& Acc.\ (K./N.) & E. Gain \\
\midrule

GPT-5
& 87.4 (89.0/65.1) & $1.00\times$
& 85.4 (87.8/63.8) & $1.00\times$
& 87.3 (88.0/56.5) & $1.00\times$ \\

BioCLIP 2
& 72.7 (74.5/47.8) & $1.00\times$
& 64.4 (67.3/36.0) & $1.00\times$
& 84.7 (85.2/60.1) & $1.00\times$ \\

SpeciesNet
& 93.7 (94.4/84.2) & $1.00\times$
& 69.2 (71.8/44.0) & $1.00\times$
& 91.6 (91.9/78.2) & $1.00\times$ \\
\midrule

On-Device
& 70.1 (75.1/0.0) & $2.68\times$
& 73.8 (81.4/0.0) & $1.48\times$
& 78.2 (80.3/0.0) & $1.18\times$ \\

CACTUS
& 66.5 (71.4/0.0) & $2.54\times$
& 74.0 (81.5/0.0) & $2.34\times$
& 78.4 (80.1/0.0) & $3.04\times$ \\

Palleon
& 69.1 (74.0/0.0) & $2.89\times$
& 73.2 (80.7/0.0) & $1.94\times$
& 76.8 (78.5/0.0) & $1.49\times$ \\

EdgeBoost
& 69.0 (74.1/0.0) & $2.33\times$
& 74.3 (81.9/0.0) & $1.33\times$
& 78.2 (80.0/0.0) & $1.27\times$ \\

EffEdge-B0
& 69.5 (74.5/0.0) & $4.30\times$
& 75.1 (82.8/0.0) & $1.70\times$
& 79.7 (81.4/0.0) & $2.91\times$ \\

EffEdge-B3
& 70.4 (75.4/0.0) & $2.78\times$
& 76.3 (84.1/0.0) & $1.62\times$
& 80.7 (82.5/0.0) & $1.32\times$ \\
\midrule

\pjn{}
& 76.6 (78.1/55.5) & $2.62\times$
& 81.5 (84.0/59.1) & $2.43\times$
& 81.6 (82.2/53.7) & $3.44\times$ \\

\pjn{}-CS
& 75.9 & $2.49\times$
& 79.0 & $2.47\times$
& 81.5 & $3.16\times$ \\
\bottomrule
\end{tabular}%
}
\end{table}

\begin{figure}[t]
    \centering
    \includegraphics[width=\columnwidth]
    {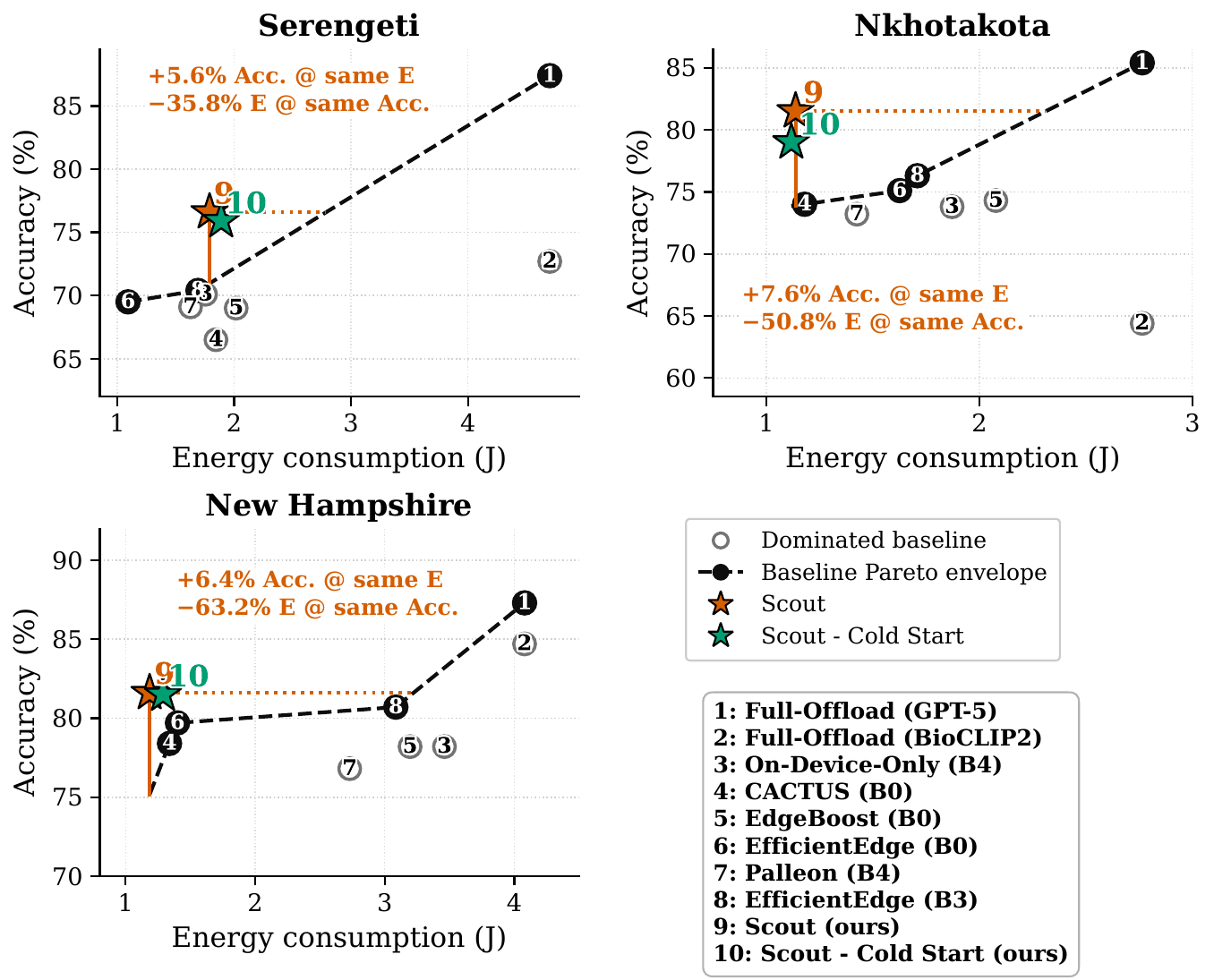}
    \caption{\textbf{Accuracy--energy trade-off.}
    The dashed line in each plot shows the Pareto envelope formed by the baselines. Both \pjn{} and \pjn{}-CS are above the baseline envelope on all three datasets.}
    \label{fig:accuracy_energy}
\end{figure}

\mypar{Overall Performance}
\pjn{} obtains 76.6\%, 81.5\%, and 81.6\% overall accuracy on Serengeti,
Nkhotakota, and New Hampshire, improving over the most accurate edge baseline
by 6.2, 5.2, and 0.9\%. On Nkhotakota and New Hampshire it
also achieves a higher energy gain than every edge baseline; on Serengeti four
baselines have higher energy gains, but each is at least 6.2\% less accurate. On known
species \pjn{} differs from the best edge baseline by only $+2.7$, $-0.1$, and
$-0.3$ points, so pruning B0 largely preserves accuracy on the initial class
set while cutting per-frame inference energy by $1.9\times$, $4.8\times$, and
$4.1\times$. The remaining accuracy gains come from species outside that set
(Section~\ref{sec:exp:novel}).

\mypar{Accuracy--Energy Trade-off}
Figure~\ref{fig:accuracy_energy} compares the methods in the
accuracy--energy space. Relative to the baseline envelope, which includes the
six edge baselines plus Full-Offload GPT-5 and BioCLIP~2, \pjn{} improves
accuracy by 5.6, 7.6, and 6.4\% at matched energy, or reduces
deployment energy by 35.8\%, 50.8\%, and 63.2\% at matched accuracy. We exclude SpeciesNet from the Pareto envelope because Snapshot Serengeti is included in its training data and potential overlap with New Hampshire is unknown. For Nkhotakota, which is outside its training distribution, SpeciesNet falls below the envelope and does not affect the comparison.

\mypar{Cold Start}
Initialized from only the deployment location and background pool, \pjn{}-CS
reaches 75.9\%, 79.0\%, and 81.5\% accuracy, within 0.7, 2.5, and 0.1\%
of \pjn{} in the controlled setting. It remains more accurate than every
controlled-setting edge baseline on all three datasets and stays above the
baseline envelope, so location-based initialization is a practical substitute for a predefined species list.

\mypar{Global and Open-Vocabulary Models}
Full-Offload (GPT-5) is the most accurate overall, but uploads every image and
therefore spends $2.62$--$3.44\times$ \pjn{}'s deployment energy. \pjn{}
outperforms BioCLIP~2 by 3.9 and 17.1 points on Serengeti and Nkhotakota; on New
Hampshire, BioCLIP~2 is 3.1\% more accurate at $3.44\times$ the energy.
SpeciesNet reaches 93.7\% on Serengeti, which is in its training data, but
drops to 69.2\% at Nkhotakota, which is not, putting it 12.3 points below
\pjn{} at $2.43\times$ the energy even though all evaluated species are in its
label set.

\subsection{Performance on Novel Species}
\label{sec:exp:novel}

We evaluate how well \pjn{} recognizes species outside the initial class set
using the novel-species results in Table~\ref{tab:end_to_end}. The full-offload
models do not use this set, so we compute their accuracy on the same image
subset only for comparison.

All six edge baselines obtain $0\%$ novel-species accuracy because their output
layers include only the initial class set. Selective offloading and
context switching can change which model processes an image, but they do not
expand the class set. In comparison, \pjn{} obtains 55.5\%,
59.1\%, and 53.7\% accuracy on novel species on Serengeti, Nkhotakota, and New
Hampshire, respectively. Full-Offload (GPT-5), which sends every image to the
same VLM and is not limited by OoC detection, obtains 65.1\%, 63.8\%, and
56.5\%. We use these results as the upper bound for \pjn{}. Across the three datasets, \pjn{} comes within 2.8--9.6\% of Full-Offload (GPT-5) while consuming 59--71\% less deployment energy. \pjn{} outperforms BioCLIP~2 by 7.7 and 23.1\% on Serengeti and Nkhotakota. On New Hampshire, BioCLIP~2 is 6.4\% more accurate but consumes
$3.44\times$ \pjn{}'s deployment energy. On Nkhotakota, which is absent from SpeciesNet's training sites, \pjn{} is 15.1\% more accurate while consuming 58.8\% less deployment energy. All evaluated species are present in the SpeciesNet label set, so this difference does not result from missing species.

An analysis of the New Hampshire dataset suggests that much of the gap to Full-Offload (GPT-5) comes from rare species that are not triggered by the OoC head.
Species discovered by \pjn{} occur 20 times on
average, compared with 3 for missed species. Rare species therefore provide
fewer opportunities for the OoC head to upload a frame and for the cloud to
identify the species and update the classifier.

\subsection{LLM Planner}
\label{sec:exp:planner}

In this section, we evaluate the performance of \pjn{}'s LLM planner by isolating the context and model selection part for \pjn{} and baselines. We compare its selections with those of rule-based methods and equal-budget random search. We make two changes from the end-to-end evaluation. First, every method is given all species that occur at that location before processing the stream, removing the advantage \pjn{} has from covering novel species. Second, when a method's OoC or switching mechanism triggers its more powerful classifier, we assume that this classifier returns the correct prediction, removing the effect of classifier choice from the comparison. Each method still runs its selected edge model and OoC head and pays for
inference, image uploads, and model downloads. The comparison therefore reflects the class context and model configuration selected by each method. All methods use
the same synthesized data and input-resolution and pruning ranges. Random
Search randomly samples the class context, input resolution, and pruning ratio,
evaluates the same number of candidates as \pjn{} on average, and uses the same
validation trace and accuracy--energy objective to select the deployed model.

\begin{table}[t]
\centering
\caption{Accuracy and energy gain for different context and model selection
methods. Random search uses the same candidate-evaluation budget per event as the LLM planner. Energy gain is relative to Full-Offload in Table~\ref{tab:end_to_end}.}
\label{tab:planner}
\scriptsize
\setlength{\tabcolsep}{3pt}
\renewcommand{\arraystretch}{1.12}
\resizebox{\columnwidth}{!}{%
\begin{tabular}{lcc cc}
\toprule
\multirow{2}{*}{Selection Method}
& \multicolumn{2}{c}{Serengeti}
& \multicolumn{2}{c}{Nkhotakota} \\
\cmidrule(lr){2-3}
\cmidrule(lr){4-5}
& Acc. (\%) & E. Gain
& Acc. (\%) & E. Gain \\
\midrule
Palleon (B4)~\cite{palleon}
& 74.2 & $2.77\times$
& 75.8 & $1.70\times$ \\
CACTUS (B0)~\cite{rastikerdar2024cactus}
& 81.8 & $2.72\times$
& 82.8 & $2.49\times$ \\
Random Search
& 81.5 & $2.97\times$
& 82.2 & $2.75\times$ \\
\midrule
\pjn{} LLM Planner (B0)
& \textbf{84.7} & $\mathbf{3.11\times}$
& \textbf{85.0} & $\mathbf{2.73\times}$ \\
\bottomrule
\end{tabular}%
}
\end{table}

Table~\ref{tab:planner} shows that the LLM planner achieves higher accuracy than the rule-based methods and Random Search while maintaining competitive energy efficiency. Compared with CACTUS, it improves accuracy by 2.9\% and 2.2\% and reduces deployment energy by 12.5\% and 8.8\% on Serengeti and Nkhotakota, respectively. Compared with Random Search, it improves accuracy by 3.2 and 2.8\%, with lower energy on Serengeti and similar energy on Nkhotakota. Both use the same average candidate budget, validation trace, and selection objective, but the LLM planner uses results from earlier searches and candidates evaluated during the current search to guide new proposals. These results show that this feedback helps the planner use its limited search budget more effectively.

\subsection{Sensitivity Analysis}
\label{sec:exp:sensitivity}

We test sensitivity to the initial class-set size and communication bandwidth on the Serengeti deployments, following the end-to-end setup in all other respects.

\begin{figure}[t]
    \centering
    \includegraphics[width=\columnwidth]
    {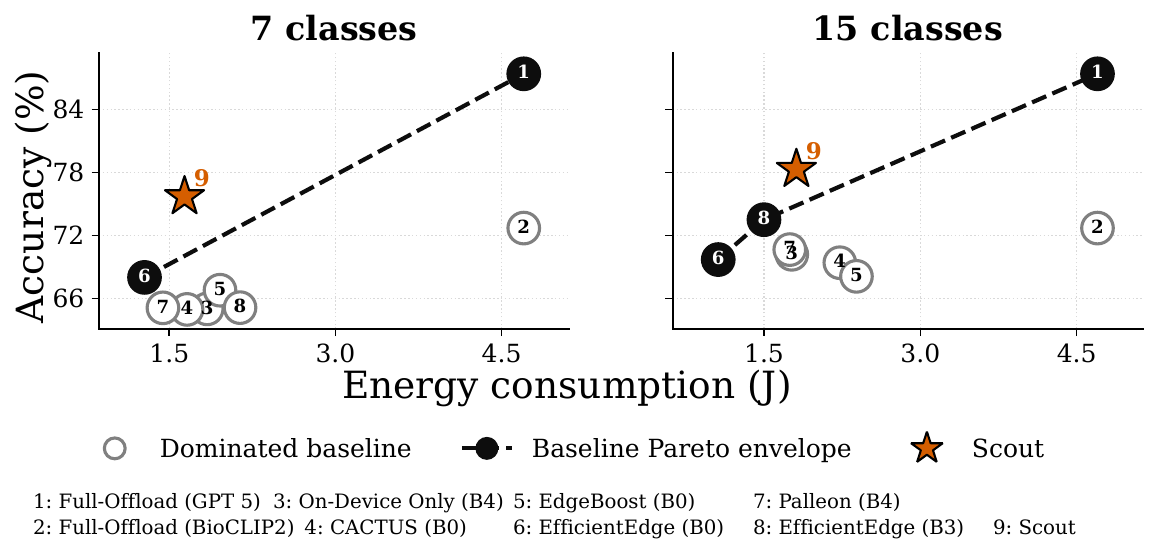}
    \caption{\textbf{Initial class-set size.}
    Accuracy and deployment energy on Serengeti when the initial class-set size is 7 or 15 species in controlled evaluation. The dashed line shows the
    baseline Pareto envelope.}
    \label{fig:initial_classes}
\end{figure}

\mypar{Initial Class-Set Size}
We repeat the controlled evaluation with 7 and 15 initial species instead of 10,
giving every method the same initial set. \pjn{} remains above the baseline
Pareto envelope in both cases (Figure~\ref{fig:initial_classes}). The margin
narrows as the initial set grows, because the fixed-label baselines then cover a
larger fraction of the stream.

\begin{figure}[t]
    \centering
    \includegraphics[width=\columnwidth]
    {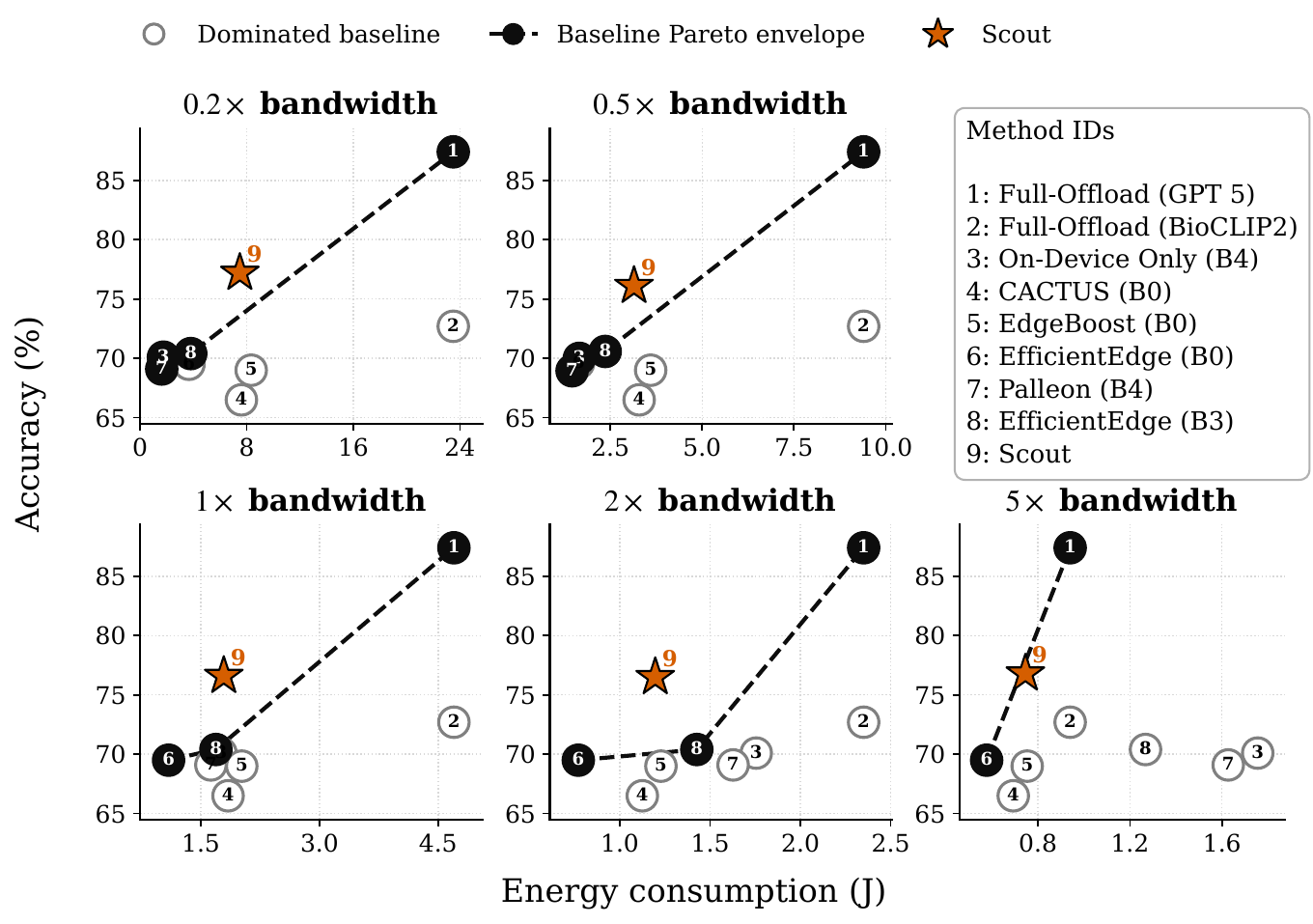}
    \caption{\textbf{Bandwidth sweep.}
    Accuracy and deployment energy on Serengeti for regional throughput scaled by $0.2$, $0.5$, $1$, $2$, and $5$. The dashed
    line shows the baseline Pareto envelope at each bandwidth.}
    \label{fig:bandwidth}
\end{figure}

\mypar{Communication Bandwidth}
Actual bandwidth varies across camera locations and over time, so we scale the
Serengeti uplink and downlink throughputs by $0.2\times$ to $5\times$ and repeat
the evaluation at each setting. \pjn{} remains above the baseline envelope from
$0.2\times$ through $2\times$ (Figure~\ref{fig:bandwidth}). At $5\times$,
uploading becomes cheap enough that Full-Offload (GPT-5) gives a slightly better
accuracy--energy trade-off. \pjn{} therefore provides its largest benefit when
communication is limited.

%% file: sec/5_limitations.tex
\section{Limitations}
\label{sec:limitations}

We evaluate \pjn{} by replaying camera-trap streams and measuring inference
energy on edge hardware, not in a long-term field deployment, so connectivity
interruptions, hardware failures, and environmental effects are not captured.
Without connectivity, \pjn{} continues using its current model but cannot update
it. Finally, the current implementation uses GPT-5 for both species
identification and planning, and the API cost of frontier models may limit the
scale of long-term deployments.

%% file: sec/6_conclusion.tex
\section{Conclusion}
\label{sec:conclusion}

We presented \pjn{}, which builds and maintains an on-device camera-trap
classifier from only the deployment location and empty frames from the site.
\pjn{} constructs site-conditioned training data, updates the classifier when
the cloud VLM identifies a new species, and uses an LLM planner to jointly
select the class context and model configuration. Across 30 camera locations,
\pjn{} improves accuracy by 5.6–7.6\% at matched energy or
reduces deployment energy by 35.8–63.2\% at matched accuracy, and obtains
53.7–59.1\% accuracy on species that fixed-label edge models cannot predict.
Initialized from the location alone, it stays within 0.1–2.5\% of these
results.

%% file: sec/7_supplementary.tex
\setcounter{table}{0}
\renewcommand{\thetable}{A\arabic{table}}

\section*{Supplementary Material}
\addcontentsline{toc}{section}{Supplementary Material}

\section{Additional Evaluation Details}
\label{sec:supp:experimental_details}

\noindent\begin{minipage}[t]{0.36\textwidth}
\mypar{Planner search budget}
At each planning event, the LLM selects up to $N$ candidate class contexts
and performs up to $K$ rounds of model-configuration search for each
context. We use $N=4$ and $K=3$ by default.
\end{minipage}\hfill
\begin{minipage}[t]{0.60\textwidth}

\mypar{Communication parameters}
As described in the main paper, we obtain regional throughput estimates
from the crowdsourced Opensignal and SpeedChecker measurement platforms.
Table~\ref{tab:supp:communication_energy} reports the resulting throughput
and transfer-energy values used in our evaluation.

\begin{center}
    \vspace{-0.4em}
    \captionof{table}{Regional communication throughput and corresponding transfer
    energy per MB used in our evaluation.}
    \label{tab:supp:communication_energy}
    \small
    \setlength{\tabcolsep}{4pt}
    \begin{tabular}{lcccc}
        \toprule
        & \multicolumn{2}{c}{Uplink}
        & \multicolumn{2}{c}{Downlink} \\
        \cmidrule(lr){2-3}
        \cmidrule(lr){4-5}
        Dataset
        & Mbps & J/MB
        & Mbps & J/MB \\
        \midrule
        Serengeti     & 5.1  & 4.7 & 7.8  & 3.1 \\
        Nkhotakota    & 9.0  & 2.7 & 17.5 & 1.4 \\
        New Hampshire & 11.9 & 2.0 & 46.5 & 0.5 \\
        \bottomrule
    \end{tabular}
\end{center}
\end{minipage}
\vspace{+0.8em}

\mypar{End-to-end energy breakdown}
Table~\ref{tab:supp:end_to_end_complete} adds total deployment energy and
its breakdown to Table~\ref{tab:end_to_end} in the main paper.
\vspace{-0.4em}
\begin{center}
\centering
\captionof{table}{Complete end-to-end results averaged over 10 camera locations per
dataset. Accuracy is reported overall and, in parentheses, for known/novel
species; \pjn{}-CS reports overall accuracy only. Total deployment energy is
in J/frame, while its inference (Inf.), upload (Up.), and download (Dl.)
components are in mJ/frame. Energy Gain is relative to Full-Offload. The
full-offload classifiers share an energy breakdown because each uploads
every frame and cloud compute energy is excluded.
Components may not sum exactly to the displayed total because of rounding.}

\label{tab:supp:end_to_end_complete}
\scriptsize
\setlength{\tabcolsep}{4pt}
\renewcommand{\arraystretch}{0.94}
\begin{tabular}{l c c c rrr}
\toprule
\multirow{2}{*}{Method}
& \multirow{2}{*}{Acc. (Known/Novel)}
& \multirow{2}{*}{Total (J/frame)}
& \multirow{2}{*}{Energy Gain}
& \multicolumn{3}{c}{Breakdown (mJ/frame)} \\
\cmidrule(lr){5-7}
& & & & Inf. & Up. & Dl. \\
\midrule
\multicolumn{7}{l}{{\small\bfseries Serengeti}} \\
\addlinespace[1pt]
Full-Offload (GPT-5)
& 87.4 (89.0/65.1) & 4.70 & $1.00\times$ & 0 & 4700 & 0 \\
Full-Offload (BioCLIP 2)
& 72.7 (74.5/47.8) & 4.70 & $1.00\times$ & 0 & 4700 & 0 \\
Full-Offload (SpeciesNet)
& 93.7 (94.4/84.2) & 4.70 & $1.00\times$ & 0 & 4700 & 0 \\
\addlinespace[1pt]
On-Device-Only (B4)
& 70.1 (75.1/0.0) & 1.76 & $2.68\times$ & 1755 & 0 & 0 \\
CACTUS (B0)
& 66.5 (71.4/0.0) & 1.84 & $2.54\times$ & 405 & 360 & 1080 \\
Palleon (B4)
& 69.1 (74.0/0.0) & 1.63 & $2.89\times$ & 1627 & 0 & 0 \\
EdgeBoost (B0)
& 69.0 (74.1/0.0) & 2.02 & $2.33\times$ & 437 & 1581 & 0 \\
EfficientEdge (B0)
& 69.5 (74.5/0.0) & 1.09 & $4.30\times$ & 448 & 644 & 0 \\
EfficientEdge (B3)
& 70.4 (75.4/0.0) & 1.69 & $2.78\times$ & 1162 & 529 & 0 \\
\pjn{} (B0)
& 76.6 (78.1/55.5) & 1.79 & $2.62\times$ & 608 & 557 & 625 \\
\pjn{}-CS (B0)
& 75.9 & 1.89 & $2.49\times$ & 659 & 532 & 698 \\
\midrule
\multicolumn{7}{l}{{\small\bfseries Nkhotakota}} \\
\addlinespace[1pt]
Full-Offload (GPT-5)
& 85.4 (87.8/63.8) & 2.76 & $1.00\times$ & 0 & 2764 & 0 \\
Full-Offload (BioCLIP 2)
& 64.4 (67.3/36.0) & 2.76 & $1.00\times$ & 0 & 2764 & 0 \\
Full-Offload (SpeciesNet)
& 69.2 (71.8/44.0) & 2.76 & $1.00\times$ & 0 & 2764 & 0 \\
\addlinespace[1pt]
On-Device-Only (B4)
& 73.8 (81.4/0.0) & 1.87 & $1.48\times$ & 1872 & 0 & 0 \\
CACTUS (B0)
& 74.0 (81.5/0.0) & 1.18 & $2.34\times$ & 433 & 230 & 516 \\
Palleon (B4)
& 73.2 (80.7/0.0) & 1.42 & $1.94\times$ & 1425 & 0 & 0 \\
EdgeBoost (B0)
& 74.3 (81.9/0.0) & 2.08 & $1.33\times$ & 460 & 1617 & 0 \\
EfficientEdge (B0)
& 75.1 (82.8/0.0) & 1.63 & $1.70\times$ & 471 & 1154 & 0 \\
EfficientEdge (B3)
& 76.3 (84.1/0.0) & 1.71 & $1.62\times$ & 1236 & 473 & 0 \\
\pjn{} (B0)
& 81.5 (84.0/59.1) & 1.14 & $2.43\times$ & 260 & 450 & 427 \\
\pjn{}-CS (B0)
& 79.0 & 1.12 & $2.47\times$ & 415 & 504 & 198 \\
\midrule
\multicolumn{7}{l}{{\small\bfseries New Hampshire}} \\
\addlinespace[1pt]
Full-Offload (GPT-5)
& 87.3 (88.0/56.5) & 4.08 & $1.00\times$ & 0 & 4080 & 0 \\
Full-Offload (BioCLIP 2)
& 84.7 (85.2/60.1) & 4.08 & $1.00\times$ & 0 & 4080 & 0 \\
Full-Offload (SpeciesNet)
& 91.6 (91.9/78.2) & 4.08 & $1.00\times$ & 0 & 4080 & 0 \\
\addlinespace[1pt]
On-Device-Only (B4)
& 78.2 (80.3/0.0) & 3.46 & $1.18\times$ & 3462 & 0 & 0 \\
CACTUS (B0)
& 78.4 (80.1/0.0) & 1.34 & $3.04\times$ & 841 & 313 & 185 \\
Palleon (B4)
& 76.8 (78.5/0.0) & 2.73 & $1.49\times$ & 2730 & 0 & 0 \\
EdgeBoost (B0)
& 78.2 (80.0/0.0) & 3.20 & $1.27\times$ & 853 & 2343 & 0 \\
EfficientEdge (B0)
& 79.7 (81.4/0.0) & 1.40 & $2.91\times$ & 887 & 514 & 0 \\
EfficientEdge (B3)
& 80.7 (82.5/0.0) & 3.09 & $1.32\times$ & 2192 & 894 & 0 \\
\pjn{} (B0)
& 81.6 (82.2/53.7) & 1.19 & $3.44\times$ & 537 & 587 & 62 \\
\pjn{}-CS (B0)
& 81.5 & 1.29 & $3.16\times$ & 603 & 604 & 82 \\
\bottomrule
\end{tabular}
\end{center}